%% file: main.tex
\documentclass[11pt]{article}

\usepackage[preprint]{acl}

\usepackage{times}
\usepackage{amsmath,amssymb}
\usepackage{latexsym}
\usepackage{booktabs}
\usepackage[table]{xcolor}
\usepackage{multirow}

\usepackage[T1]{fontenc}

\usepackage[utf8]{inputenc}

\usepackage{microtype}

\usepackage{inconsolata}

\usepackage{graphicx}

\title{Physio-DPO: Aligning Large Language Models with the Protein Energy Landscape to Eliminate Structural Hallucinations}

\author{
Qiwei Meng \\
Xi'an Jiaotong University \\
\texttt{Twilight\_M@stu.xjtu.edu.cn}
}

\begin{document}
\maketitle
\begin{abstract}
Large Protein Language Models have shown strong potential for generative protein design, yet they frequently produce structural hallucinations, generating sequences with high linguistic likelihood that fold into thermodynamically unstable conformations. Existing alignment approaches such as Direct Preference Optimization are limited in this setting, as they model preferences as binary labels and ignore the continuous structure of the physical energy landscape. We propose Physio-DPO, a physics informed alignment framework that grounds protein language models in thermodynamic stability. Physio-DPO introduces a magnitude aware objective that scales optimization updates according to the energy gap between native structures and physics perturbed hard negatives. Experiments show that Physio-DPO consistently outperforms strong baselines including SFT, PPO, and standard DPO, reducing self consistency RMSD to 1.28 \AA\ and increasing foldability to 92.8\%. Qualitative analysis further demonstrates that Physio-DPO effectively mitigates structural hallucinations by recovering biophysical interactions such as hydrophobic core packing and hydrogen bond networks.
\end{abstract}

\section{Introduction}
Recent advances in scaling Protein Language Models (PLMs), exemplified by ESM-series \citep{lin2023evolutionary, hayes2024simulating} and ProGen \citep{madani2023large}, have substantially advanced computational protein design. By internalizing the statistical grammar of evolution from billions of sequences, these models exhibit strong generative capabilities and can produce protein-like sequences \textit{de novo}. However, a fundamental misalignment remains. The training objective of PLMs, minimizing token-level perplexity, serves only as an indirect proxy for evolutionary fitness and does not explicitly optimize thermodynamic stability. As a result, even large-scale PLMs frequently produce \textit{structural hallucinations}: sequences in which the model expresses high confidence, yet which fold into high-energy, physically invalid conformations characterized by disordered regions, steric clashes, or exposed hydrophobic cores \citep{anishchenko2021novo,gopalan2025hallucinations}.

Direct Preference Optimization (DPO) \citep{rafailov2023direct} has recently emerged as a more stable, offline alternative by reformulating reinforcement learning as a classification objective over preference pairs. While DPO has proven effective for alignment~\citep{das2025dpo}, its standard formulation is ill-suited for biophysical optimization. DPO models preferences as binary relations, discarding the magnitude of quality differences between candidates. In physical systems, however, the energy gap between a native structure and a decoy encodes essential information about the topology and steepness of the energy landscape. Collapsing continuous thermodynamic signals into binary labels prevents the model from distinguishing minor fluctuations from severe structural failures. Additionally, reference-free variants such as SimPO \citep{meng2024simpo}, ORPO \citep{hong2024orpo}, which remove explicit regularization, risk eroding the evolutionary priors that underpin biological plausibility.

To address this mismatch between discrete preference learning and continuous physical laws, we propose \textbf{Physio-DPO}, a physics-informed alignment framework designed to ground large-scale PLMs in thermodynamic reality. Physio-DPO extends standard DPO with a magnitude-aware objective that explicitly weights optimization updates according to the physical energy gap, enabling the model to focus its capacity on resolving substantial stability barriers. We further introduce a hard negative mining strategy that generates adversarial decoys which are linguistically plausible yet structurally unsound, forcing the model to learn fine-grained biophysical distinctions. Extensive experiments on protein generation demonstrate that physics-informed preference optimization can achieve stable, scalable, and physically grounded protein generation at unprecedented model scales. 

Our contributions are summarized as follows: \textbf{(1)} We introduce a large-scale physics-grounded preference dataset containing 1M native--decoy pairs, where hard negatives are generated via targeted physical perturbations to expose subtle yet critical structural failures. \textbf{(2)} We propose a physics-informed preference optimization framework that extends standard DPO with a continuous, magnitude-aware objective, enabling gradient updates to scale with thermodynamic energy gaps. \textbf{(3)} We provide a theoretical analysis showing that the proposed energy-weighted objective reduces gradient variance and corresponds to optimizing a principled surrogate of the underlying physical energy distribution. \textbf{(4)} Extensive experiments on large-scale protein generation demonstrate that Physio-DPO consistently outperforms strong baselines, achieving state-of-the-art structural accuracy (sc-RMSD of \textbf{1.28 \AA}) while substantially mitigating structural hallucinations.

\section{Related Work}

\paragraph{Generative Protein Models and Hallucinations.} Early approaches to protein generation relied on statistical correlations derived from Multiple Sequence Alignments (MSAs), such as Potts models \citep{levy2017potts}. The scaling of Transformer architectures \citep{vaswani2017attention} has shifted the paradigm towards auto-regressive PLMs trained on massive metagenomic databases \citep{rives2021biological, elnaggar2021prottrans}. Models like ESM-series \citep{,hayes2024simulating, lin2023evolutionary} and ProtGPT2 \citep{ferruz2022protgpt2} capture long-range evolutionary dependencies, enabling the generation of diverse sequences. However, these models optimize a token-level cross-entropy loss, which is a proxy for evolutionary fitness but not a direct measure of structural stability. As a result, they are prone to \textit{hallucinations}: sequences that appear statistically plausible but fail to fold into defined tertiary structures due to steric clashes or unsatisfied hydrogen bonds \citep{anishchenko2021novo}. While diffusion models \citep{watson2023novo, ingraham2023illuminating, lisanza2025multistate} explicitly generate structure, they are computationally expensive and lack the sequence-design flexibility of PLMs. Our work retains the efficiency of PLMs while enforcing structural validity through alignment.

\paragraph{Alignment.} Aligning language models to specific objectives has traditionally relied on RLHF~\citep{zhou2025sequence,mei2025real}, most commonly implemented with Proximal Policy Optimization (PPO)\citep{schulman2017proximal}. In protein design, reinforcement learning has been applied to optimize properties such as solubility or fluorescence\citep{angermueller2019model}; however, PPO requires training separate reward and value networks, often resulting in instability and high memory overhead. Direct Preference Optimization (DPO)\citep{rafailov2023direct} provides a more stable, offline alternative by recasting RL as preference-based classification. Subsequent NLP-focused extensions, including IPO\citep{azar2024general} and KTO~\citep{ethayarajh2024kto}, refine margin handling, while reference-free variants such as SimPO~\citep{meng2024simpo} and ORPO~\citep{hong2024orpo} improve efficiency. Nevertheless, these approaches remain suboptimal for biological sequence design: reference-free methods discard KL regularization, risking catastrophic forgetting of evolutionary priors, and all treat preferences as binary labels, ignoring the magnitude of physical signals. \textbf{Physio-DPO} overcomes these limitations by retaining KL anchoring to preserve biological plausibility while incorporating thermodynamic magnitudes directly into preference optimization, bridging semantic alignment and physical validity.

\begin{figure*}[tbh]
    \centering
    \includegraphics[width=0.935\textwidth]{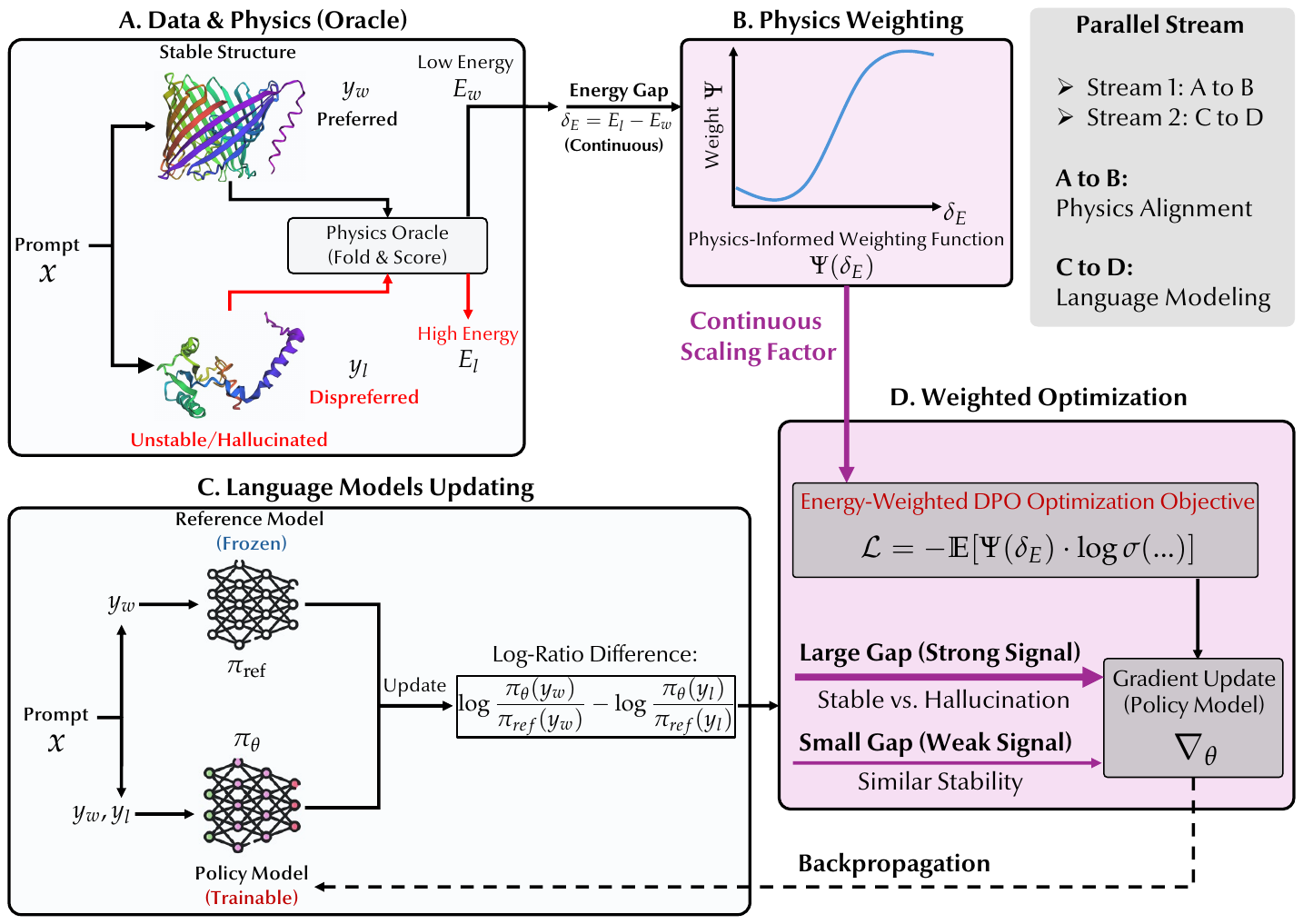}
    \caption{Ovewview of Physio-DPO framework. The physics stream folds a sampled pair $(y_w, y_l)$ and computes an energy gap $\delta_E$, which is mapped to a weight $\Psi(\delta_E)$. The language modeling stream computes the DPO log ratio using the policy $\pi_\theta$ and a frozen reference $\pi_{\mathrm{ref}}$. Physio-DPO reweights each DPO term by $\Psi(\delta_E)$, amplifying updates from pairs with large stability gaps and aligning the model with a continuous energy landscape.}
    \label{fig:overview}
    \vspace{-12pt}
\end{figure*}

\section{Preliminaries}
\paragraph{Direct Preference Optimization (DPO).}
Direct Preference Optimization (DPO)~\citep{rafailov2023direct} is a framework for aligning language models with preference data without reinforcement learning. Given a reference policy $\pi_{\text{ref}}$ and a trainable policy $\pi_\theta$, it optimizes $\pi_\theta$ to prefer a winner $y_w$ over a loser $y_l$ by maximizing the probability:
\begin{equation}
    P(y_w \succ y_l) = \sigma\!\left(r(x,y_w) - r(x,y_l)\right),
\end{equation}
where $\sigma$ is the sigmoid and $r(x,y)$ follows the Bradley–Terry model, yielding a stable preference-based objective regularized by $\pi_{\text{ref}}$.

\paragraph{Problem Formulation.}
We formulate protein design as an unconditional sequence generation problem. Let $x$ denote a generic prefix (e.g., a start token); an autoregressive PLM parameterized by $\theta$ defines a policy $\pi_\theta(y|x)$ over amino acid sequences $y \in \mathcal{Y}$. While pretrained PLMs capture evolutionary plausibility, they do not explicitly enforce biophysical validity. We assume access to a physical energy oracle $\mathcal{E}(y)$, where lower energy is higher thermodynamic stability. Our objective is to align $\pi_\theta$ to favor low-energy sequences while remaining close to a reference model $\pi_{\mathrm{ref}}$ to keep diversity.

\section{Methodology}

\begin{figure*}[tbh]
  \centering
  \includegraphics[width=.9\textwidth]{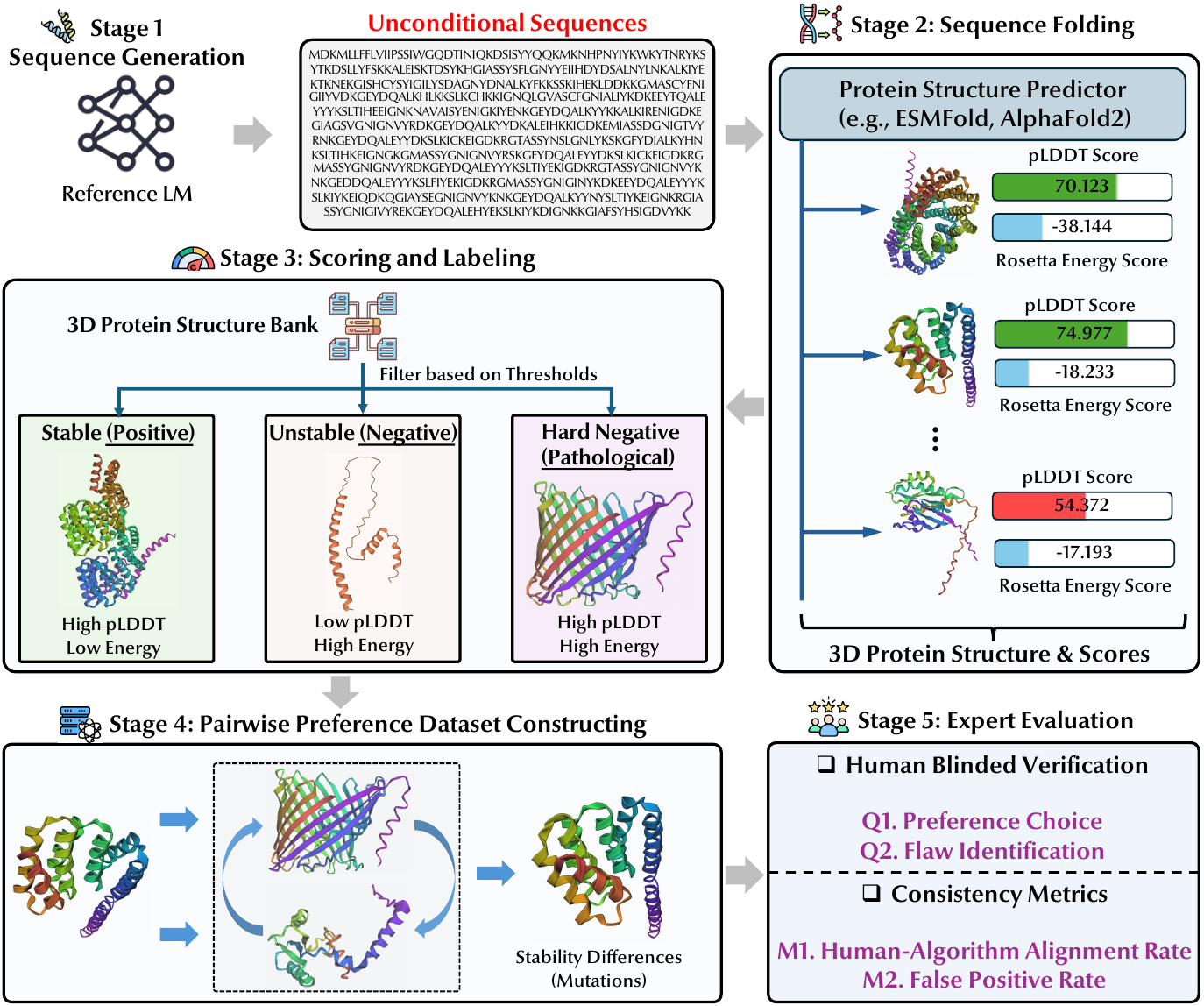}
  \caption{Construction pipeline for PhysioPref-1M. \textbf{Step 1:} diverse sequence generation from a reference language model~\citep{ferruz2022protgpt2}; \textbf{Step 2:} structure prediction via folding; \textbf{Step 3:} scoring and labeling based on pLDDT confidence~\citep{fang2025alphafold} and Rosetta energy scores~\citep{alford2017rosetta}, including the identification of hard negatives with high confidence but poor stability; \textbf{Step 4:} construction of preference pairs that maximize stability gaps; and \textbf{Step 5:} human-in-the-loop evaluation to verify alignment between labels and biophysical judgment.}
  \label{fig:dataset}
  \vspace{-12pt}
\end{figure*}

While standard DPO provides a stable alignment objective, it treats preferences as binary and ignores magnitude information~\citep{liu2025survey}. In biophysical settings, where energy gaps reflect structural instability, this discretization discards essential information from the continuous energy landscape. We therefore propose \textbf{Physio-DPO}, a physics-informed alignment framework that incorporates thermodynamic energy magnitudes into preference optimization. As shown in Fig.~\ref{fig:overview}, Physio-DPO comprises two stages: (i) constructing a physics-grounded preference dataset via a Generate–Fold–Score pipeline (Sec.\ref{sec:datasets}); and (ii) optimizing an energy-weighted objective that scales DPO gradients by physical stability gaps (Sec.~\ref{sec:physio}).

\subsection{PhysioPref-1M Benchmark}\label{sec:datasets}
Effective preference alignment in protein design requires dense, physically grounded supervision, which is largely absent from existing instruction-tuning datasets. To this end, we introduce PhysioPref-1M, a large-scale preference benchmark comprising 1M protein pairs annotated by thermodynamic criteria. The dataset is constructed via an adversarial generation and filtration pipeline (Fig.~\ref{fig:dataset}) that deliberately induces and identifies hard negatives—structures that appear foldable yet violate physical stability. Preference pairs are formed by contrasting stable proteins against unstable or pathological decoys, ensuring informative energy gaps. A human-in-the loop evaluation further validates the reliability of the automated labeling.

\subsection{Optimization Objective}\label{sec:physio}
Let $\mathcal{E}(y)$ denote the physical energy. We assume that preference strength is continuous rather than binary, and is governed by a Boltzmann distribution over energy differences. For $(y_w, y_l)$, the preference strength is determined by the energy gap:
\begin{equation}
\delta_E(y_w, y_l) = \text{ReLU}\big(\mathcal{E}(y_l) - \mathcal{E}(y_w)\big)
\end{equation}
Standard DPO maximizes the log-likelihood of preferred responses relative to a reference. We extend this objective with a magnitude-aware formulation by introducing a physics-informed weighting function $\Psi: \mathbb{R}^+ \rightarrow [0, \lambda_{\text{max}}]$, which maps the energy gap to optimization intensity.
\begin{equation}
\Psi(\delta_E) = \lambda \cdot \sigma\left(\frac{\delta_E - \mu}{\tau}\right)
\end{equation}
where $\mu$ sets the sensitivity around the critical energy boundary and $\tau$ controls the transition sharpness, suppressing noise from small $\delta_E$ while amplifying signals from hard negatives.

\paragraph{The Energy-Weighted Objective.}
We integrate $\Psi(\delta_E)$ into the DPO formulation. The Physio-DPO objective function is defined as:
\begin{equation}
\begin{aligned}
\mathcal{L}_{\text{Physio}}(\pi_\theta; \pi_{\text{ref}})
=\\ - \mathbb{E}_{(x, y_w, y_l) \sim \mathcal{D}} \Big[
& \Psi(\delta_E) \cdot \log \sigma \Big(
  \beta \log \tfrac{\pi_\theta(y_w|x)}{\pi_{\text{ref}}(y_w|x)} \\
& \qquad\qquad
- \beta \log \tfrac{\pi_\theta(y_l|x)}{\pi_{\text{ref}}(y_l|x)}
\Big)
\Big].
\end{aligned}
\end{equation}
This can be interpreted as a Cost-Sensitive Learning approach where the misclassification cost is dynamic and determined by the laws of physics.

\subsection{Gradient Modulation Analysis}
To analyze how Physio-DPO improves stability, we examine its gradient dynamics. Let
$r_\theta(x, y) = \beta \log \frac{\pi_\theta(y|x)}{\pi_{\text{ref}}(y|x)}$ denote the implicit reward. The gradient of the Physio-DPO objective is:
\begin{equation}
\nabla_\theta \mathcal{L}_{\text{Physio}}
= - \mathbb{E}\!\left[ \Psi(\delta_E)\, \sigma(-\Delta r_\theta)\, \nabla_\theta \Delta r_\theta \right],
\end{equation}
where $\Delta r_\theta = r_\theta(x, y_w) - r_\theta(x, y_l)$. This decomposition highlights a gradient modulation mechanism: the standard DPO error term $\sigma(-\Delta r_\theta)$ vanishes once preferences are confidently learned, while the physics-informed gain $\Psi(\delta_E)$ scales updates by the energy gap. As a result, gradients are suppressed for ambiguous pairs with negligible stability differences and amplified for hard negatives with large physical violations, focusing optimization on the most critical biophysical errors.

\paragraph{Theoretical Insights.}
We provide a theoretical analysis showing that Physio-DPO induces a physics-informed optimization curriculum. In the early training regime, the gradient of Physio-DPO is locally equivalent to maximizing a reward function proportional to the physical energy gap, aligning the implicit reward with the negative energy landscape. More generally, the energy-dependent weighting term provably amplifies gradient updates for hard negative pairs with severe physical violations, while suppressing updates for physically ambiguous cases. Finally, we establish a connection between Physio-DPO and thermodynamic equilibrium, showing that the objective approximates KL minimization toward a Boltzmann distribution defined by physical energy.

\section{Main Results}
We evaluate Physio-DPO on protein generation to address three questions: \textbf{(Q1)} whether incorporating physical energy landscapes improves over binary preference alignment; \textbf{(Q2)} whether Physio-DPO mitigates structural hallucinations characterized by high confidence but low stability; and \textbf{(Q3)} how the energy-weighted objective $\Psi(\delta_E)$ and hard-negative mining contribute to performance.

\subsection{Experimental Setup}

\paragraph{Datasets.} We utilize our constructed PhysioPref-1M benchmark (Section 3.2). We strictly split the dataset by sequence identity (using MMseqs2) into Train (900,000), Validation (50,000), and Test (50,000) sets, ensuring no test sequence shares $>30\%$ identity with training samples to evaluate generalization rather than memorization.

\paragraph{Baseline.}
We evaluate Physio-DPO against a set of baselines spanning different scales and alignment paradigms, including \textbf{(1) Unaligned PLMs}: ProGen2-XL (6.4B), ESM-3 Open (1.4B), and ProtGPT2 (762M); \textbf{(2) Supervised Fine-Tuning (SFT)}: ProGen2-XL fine-tuned on stable ($y_w$) subset; \textbf{(3) Reinforcement Learning (PPO)}: We apply standard RLHF using PPO, where a reward model is trained on PhysioPref-1M preference pairs; \textbf{(4) Preference Optimization Methods}: binary DPO, IPO, and KTO.

\paragraph{Implementation Details.} All models are initialized from ProGen2-XL~\citep{nijkamp2023progen2}. We utilize LoRA~\citep{hu2022lora} with rank $r=16$ and $\alpha=32$ for fine-tuning. Experiments are conducted on 4 $\times$ NVIDIA A100 (80GB) GPUs using the HuggingFace TRL. We set the DPO coefficient $\beta=0.1$. We employ the physics-informed weighting $\Psi(\delta_E)$ with scaling parameters $\mu=50$ and $\tau=10$.

\paragraph{Evaluation Metrics.}We evaluate generated proteins along four dimensions: \textbf{(1)} \textbf{structural stability}, measured by self-consistency RMSD (sc-RMSD); \textbf{(2)} \textbf{foldability}, defined as the fraction of sequences with predicted pLDDT greater than 70; \textbf{(3)} \textbf{biophysical validity}, quantified by average Rosetta energy per residue; and \textbf{(4)} \textbf{diversity}, assessed using language-model perplexity and maximum sequence identity to the training set.

\input{tables/alignment}

\subsection{Structural and Biophysical Alignment}
Table~\ref{tab:main_results} reports results on generation of 30K novel proteins. Among unaligned backbones, ProGen2-XL provides the strongest baseline with 52.4\% foldability, yet nearly half of the sequences remain non-foldable, indicating persistent structural hallucinations. Alignment markedly improves generation quality: supervised fine-tuning increases foldability to 71.5\%, while preference-based methods (DPO, IPO, KTO) further exceed 80\%. Although PPO achieves competitive structural metrics, it shows clear instability, reflected by elevated perplexity. In contrast, Physio-DPO delivers the best overall performance, reducing sc-RMSD by 0.54 \AA, improving foldability to 92.8\%, and attaining the lowest average energy (-3.05 REU) while preserving linguistic diversity. These results demonstrate that explicitly optimizing the continuous energy landscape effectively suppresses hard negatives overlooked by binary preference objectives.

\begin{figure*}[tbh]
    \centering
    \includegraphics[width=.975\textwidth]{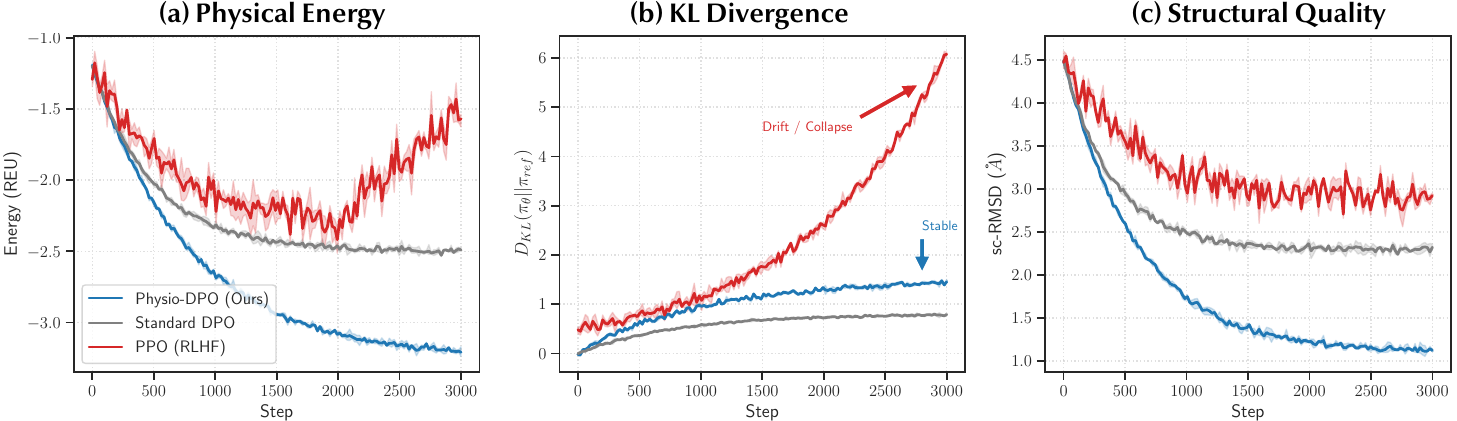}
    \caption{Training dynamics curves, (a) Physical Energy, (b) KL Divergence, (c) sc-RMSD.}
    \label{fig:dynamics}
    \vspace{-12pt}
\end{figure*}

\subsection{Training Dynamic Analysis}
To evaluate optimization efficiency and stability, we analyze training dynamics in training, tracking physical energy, KL divergence, and sc-RMSD. As shown in Fig.~\ref{fig:dynamics}(a,b), PPO becomes unstable after approximately 2K steps, with rapidly increasing KL divergence and degraded energy, indicating policy drift from reward exploitation. Standard DPO converges quickly but saturates early, limiting further improvement. In contrast, Physio-DPO continues to improve throughout training, achieving the lowest energy and sc-RMSD while maintaining a bounded KL divergence (about 1.5 nats). This demonstrates that the energy-weighted objective effectively regularizes optimization and preserves alignment with the pre-trained backbone.

\subsection{Ablation Studies}
Table~\ref{tab:ablation} shows the ablation results for Physio-DPO. Replacing hard negative mining with random negatives causes the largest performance drop, increasing sc-RMSD from 1.28 to 2.12~\AA, indicating that random negatives provide weak and uninformative gradients. Removing the physics-informed weighting term (reducing to standard DPO) further lowers foldability by 9.2\%, confirming that equal treatment of preference pairs fails to reflect the severity of physical violations. Finally, replacing the sigmoid weighting with a linear scheme degrades performance (1.45~\AA), suggesting that unbounded linear scaling is overly sensitive to extreme energy gaps, whereas the sigmoid function yields more stable and effective gradient modulation.

\begin{table}[tbh]
\centering
\resizebox{.48\textwidth}{!}{
\begin{tabular}{l|c|cc}
\toprule
\textbf{Ablation} & \textbf{Change} & \textbf{sc-RMSD} $\downarrow$ & \textbf{Foldability} $\uparrow$ \\
\midrule
\textbf{Physio-DPO} & \textit{Sigmoid + Hard Negatives} & \textbf{1.28} & \textbf{92.8\%} \\
\midrule
\emph{w/o} Weighting & Standard DPO & 1.82 & 83.6\% \\
\emph{w/o} Hard Negatives & Random Negatives & 2.12 & 76.5\% \\
\emph{w/} Linear Weighting & $\Psi(\delta_E) \propto \delta_E$ & 1.45 & 89.1\% \\
\bottomrule
\end{tabular}
}
\caption{Ablation study results. \emph{w/o}: without.}
\label{tab:ablation}
\vspace{-12pt}
\end{table}

\subsection{Mitigating Hallucinations}
\begin{figure}[tbh]
\centering
\includegraphics[width=.48\textwidth]{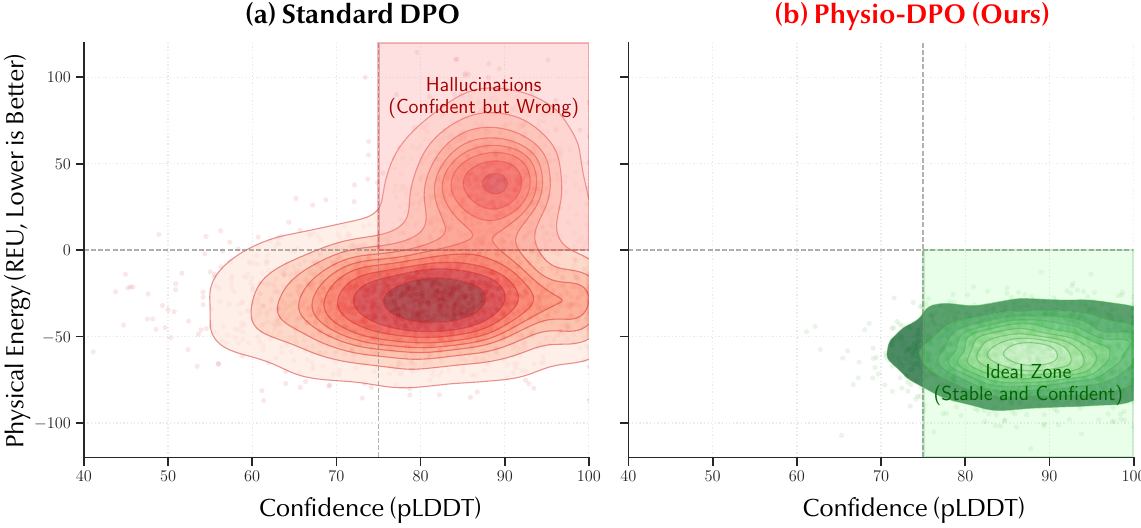}
\caption{Energy vs. Confidence (pLDDT) plane.}
\label{fig:energy_vs_confidence}
\vspace{-6pt}
\end{figure}

A critical failure mode of PLMs is generating Hallucinations-sequences that the model is confident in (low perplexity) but are biophysically invalid. We visualize the distribution of generated sequences in the Energy vs. Confidence plane (Figure \ref{fig:energy_vs_confidence}). As expected, Standard DPO shows a cluster of samples in the "High Confidence, High Energy" quadrant. These are the hallucinations. Physio-DPO successfully clears this quadrant. The shifts towards "High Confidence, Low Energy" quadrant, demonstrating that Physio-DPO effectively aligns the model's confidence with physical reality.

\begin{figure*}[tbh]
\centering
\includegraphics[width=1\textwidth]{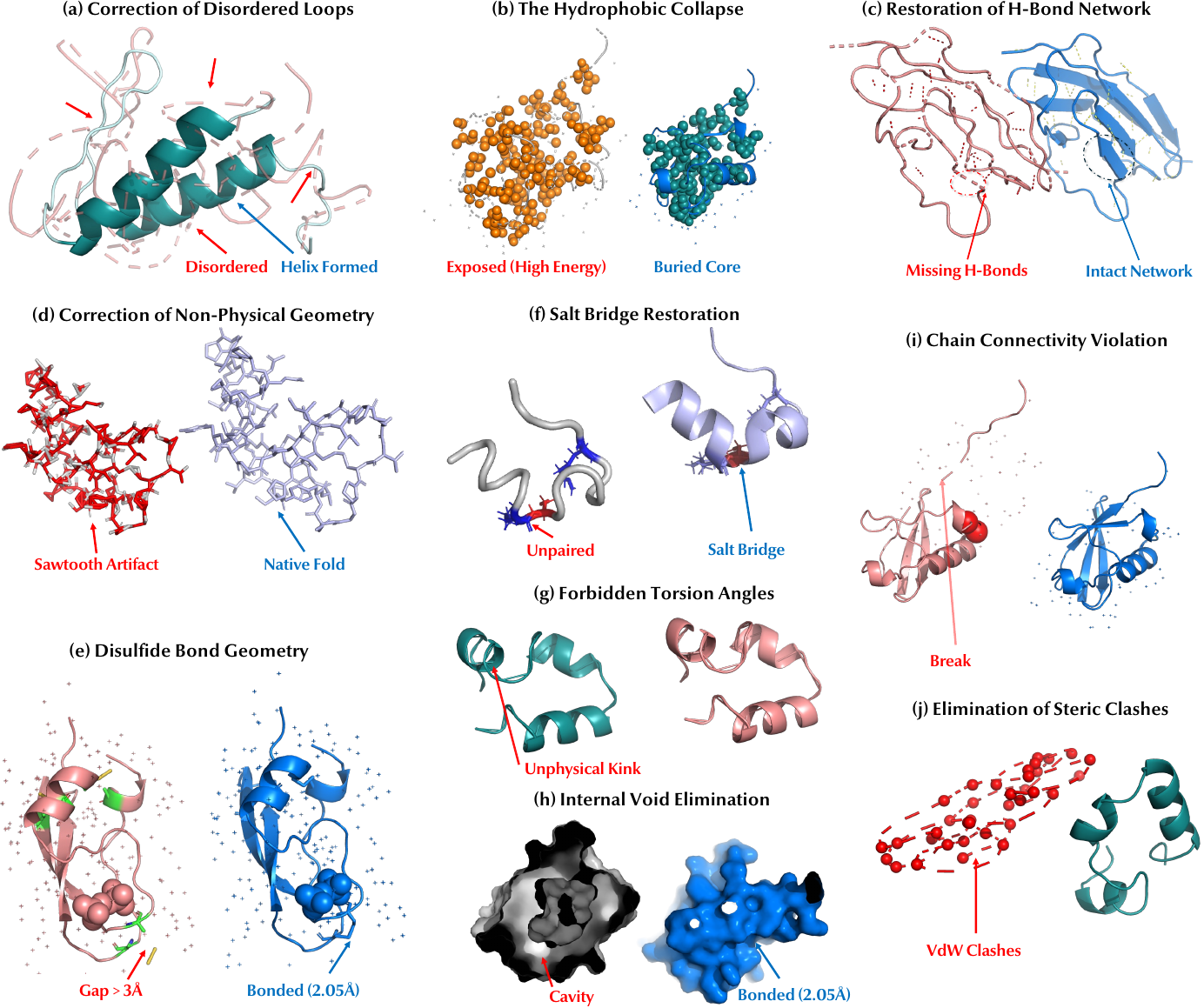}
\caption{Comprehensive qualitative analysis of biophysical validity. We compare structures generated by the SFT Baseline (\textcolor{red}{Red}/\textcolor{pink}{Pink}/\textcolor{gray}{Grey}, Left) and Physio-DPO (\textcolor{blue}{Blue}/\textcolor{teal}{Teal}, Right). \textbf{(a)} Physio-DPO compacts disordered loops into stable helices. \textbf{(b)} Exposed hydrophobic residues (\textcolor{orange}{Orange}) in SFT are buried into a tight core (\textcolor{teal}{Teal}) by Physio-DPO. \textbf{(c, e, f)} Restoration of critical atomic interactions: hydrogen bond networks in $\beta$-sheets, precise disulfide bond geometry ($2.05\text{\AA}$), and electrostatic salt bridges. \textbf{(d, g, i)} Correction of severe geometrical violations, including non-physical "sawtooth" backbones, forbidden torsion kinks, and chain connectivity breaks. \textbf{(h, j)} Optimization of packing density by eliminating destabilizing internal voids and steric clashes (\textcolor{red}{Red}).}
\label{fig:case_studies}
\vspace{-12pt}
\end{figure*}

\subsection{Structural Corrections Analysis}
To examine how Physio-DPO improves biophysical validity, Fig.~\ref{fig:case_studies} presents a visual comparison covering various structural failure modes. While the SFT baseline often preserves global topology, it frequently violates fine grained physical constraints.

\paragraph{Secondary \& Tertiary Stability.}
Physio-DPO consistently improves conformational stability by compacting disordered loop regions into well formed helices (Fig. \ref{fig:case_studies}a) and promoting hydrophobic core formation (Fig. \ref{fig:case_studies}b), thereby reducing solvation energy. It further eliminates internal voids observed in baseline structures (Fig. \ref{fig:case_studies}h), resulting in packing densities closer to native proteins.

\paragraph{Atomic Interaction Recovery.}
Physio-DPO restores key atomic interactions that are frequently disrupted in baseline generations. This includes recovering hydrogen bond networks in beta sheet regions (Fig. \ref{fig:case_studies}c), enforcing correct disulfide bond geometry (Fig. \ref{fig:case_studies}e), and pairing oppositely charged residues to form stabilizing salt bridges (Fig. \ref{fig:case_studies}f).

\paragraph{Correction of Geometrical Violations.}
The energy weighted objective also suppresses severe stereochemical violations. Compared to the baseline, Physio-DPO corrects non physical backbone distortions (Fig. \ref{fig:case_studies}d), forbidden torsion angle configurations (Fig. \ref{fig:case_studies}g), and chain connectivity breaks (Fig. \ref{fig:case_studies}i). In addition, steric clashes are substantially reduced (Fig. \ref{fig:case_studies}j), ensuring that generated structures respect Van der Waals constraints.

\input{tables/zeroshot.tex}
\subsection{Zero-shot Generalization}
To evaluate generalization beyond the synthetic distribution, we assess zero-shot performance on ProteinGym~\citep{notin2023proteingym} using log-likelihood under $\pi_\theta$ across five representative assays (Table~\ref{tab:proteingym}). Physio-DPO achieves the highest average Spearman correlation, demonstrating improved functional predictivity from physical alignment. Gains are most pronounced on stability-driven tasks (GFP and P53), consistent with effective encoding of thermodynamic constraints. In contrast, retrieval-augmented baselines remain superior on GB1, an antibody-binding task, reflecting the monomeric focus of our physics oracle. Notably, PPO degrades zero-shot performance, whereas Physio-DPO preserves pretrained semantic structure.

\subsection{Hyperparameter Sensitivity}
We evaluate the robustness of Physio-DPO with respect to two key hyperparameters: the KL-penalty coefficient ($\beta$) and the physics weighting scale ($\mu$).

\begin{figure}[tbh]
    \centering
    \includegraphics[width=.48\textwidth]{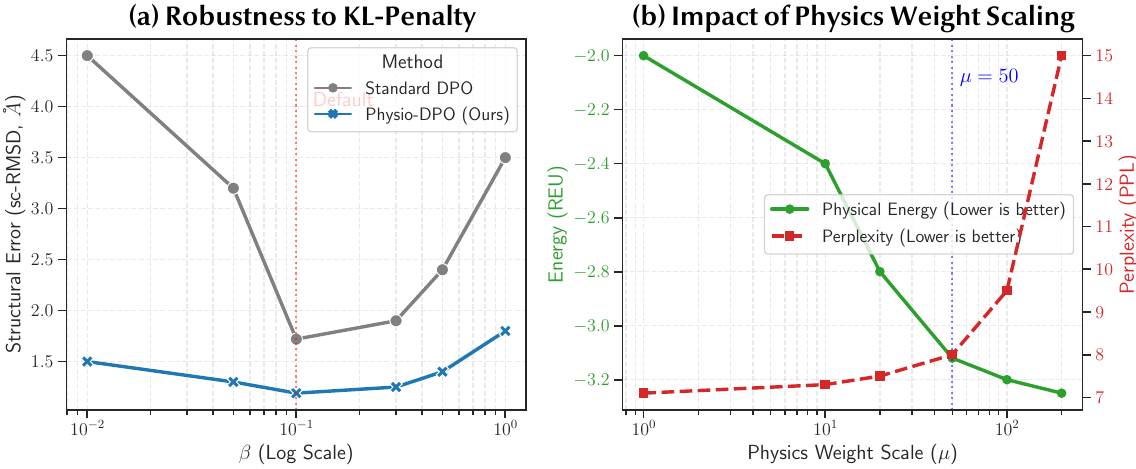}
    \caption{Hyperparameter sensitivity analysis results.}
    \label{fig:sensitivity}
    \vspace{-6pt}
\end{figure}

\paragraph{Robustness to KL Penalty.}
Figure~\ref{fig:sensitivity}(a) compares sc-RMSD for Physio-DPO and DPO across $\beta \in [0.01, 1.0]$. Physio-DPO remains stable and achieves consistently low sc-RMSD even at small $\beta$ values ($\beta=0.01$), indicating that dense, physics-informed supervision via $\Psi(\delta_E)$ effectively regularizes training and mitigates catastrophic forgetting.

\paragraph{Effect of Physics Weighting Scale.}
The $\mu$ governs the strength of energy-dependent modulation. As shown in Figure~\ref{fig:sensitivity}(b), small values of $\mu$ ($<10$) yield SFT-like behavior with higher energy, while excessively large values ($>100$) overemphasize physical energy and degrade language modeling performance, reflected by increased perplexity. We identify a broad optimal range of $\mu \in [20, 50]$, where Physio-DPO achieves low physical energy while preserving strong generative quality.

\subsection{Additional Analysis}
We further analyze the scaling behavior, stereochemical validity, and length robustness of Physio-DPO. Results show improved scaling with model size, recovery of valid Ramachandran distributions, and consistent structural quality for long sequences.

\section{Conclusion}
In conclusion, we propose Physio-DPO, a physics-informed preference optimization framework for aligning large protein language models with thermodynamic stability. We show that discrete preference modeling is insufficient in biophysical settings, as it neglects the continuous structure of energy landscapes. By incorporating energy magnitudes directly into the alignment objective, Physio-DPO guides optimization toward physically meaningful distinctions. Our results demonstrate that embedding physical principles at the alignment stage enables large-scale protein language models to internalize fine-grained biophysical constraints without sacrificing generative or linguistic capacity.

\clearpage

\section*{Limitations}
While Physio-DPO effectively aligns protein language models with thermodynamic stability, our current study focuses on monomeric folding energy as the primary physical signal. Consequently, properties involving multi-state equilibria or intermolecular interactions are not explicitly optimized. Moreover, we rely on fast physics-based oracles as approximations of true biophysical energetics, a common and practical trade-off in protein design. Notably, Physio-DPO is agnostic to the choice of energy model and can readily incorporate richer or task-specific physical signals as they become available.


\bibliography{main}
\end{document}

%% file: tables/alignment.tex
\begin{table*}[tbh]
\centering
\resizebox{.925\textwidth}{!}{
\begin{tabular}{l|ccc|c|cc}
\toprule
\multirow{2}{*}{\textbf{Method}} & \multicolumn{3}{c|}{\textbf{Structural Metrics}} & \textbf{Biophysical} & \multicolumn{2}{c}{\textbf{Diversity Metrics}} \\
& sc-RMSD (\AA) $\downarrow$ & pLDDT $\uparrow$ & Foldability (\%) $\uparrow$ & Energy (REU) $\downarrow$ & PPL $\downarrow$ & Seq-Id (\%) $\downarrow$ \\
\midrule
\multicolumn{7}{l}{\textit{Pre-trained Backbones (Zero-shot Generation)}} \\
ProtGPT2 (762M)~\citep{ferruz2022protgpt2} & 5.12 & 48.5 & 22.1 & -0.85 & 8.2 & - \\
ESM-2 (3B)$^\dagger$~\citep{lin2023evolutionary} & 4.55 & 56.2 & 35.4 & -1.05 & 6.8 & - \\
ESM-3 (1.4B)~\citep{hayes2024simulating} & 3.95 & 62.5 & 45.8 & -1.31 & 6.5 & - \\
\textbf{ProGen2-XL (6.4B)}~\citep{nijkamp2023progen2} & 3.25 & 67.8 & 52.4 & -1.65 & \textbf{6.1} & - \\
\midrule
\multicolumn{7}{l}{\textit{Alignment Methods (Backbone: ProGen2-XL 6.4B + LoRA)}} \\
\textbf{SFT} (Supervised Fine-tuning) & 2.35 & 75.2 & 71.5 & -2.25 & 7.4 & 34.1 \\
PPO (RLHF)~\citep{schulman2017proximal} & 2.15 & 78.5 & 79.2 & -2.48 & 12.5 & 39.8 \\
DPO (Standard)~\citep{rafailov2023direct} & 1.82 & 81.3 & 83.6 & -2.65 & 8.6 & 36.5 \\
IPO~\citep{azar2024general} & 1.88 & 80.8 & 82.9 & -2.58 & \underline{7.8} & 35.2 \\
KTO~\citep{ethayarajh2024kto} & 1.79 & 82.1 & 84.1 & -2.71 & 8.1 & 36.1 \\
\midrule
\textbf{Physio-DPO (Ours)} & \textbf{1.28} & \textbf{87.5} & \textbf{92.8} & \textbf{-3.05} & 8.2 & \textbf{33.8} \\
\textit{\small improvement vs. Standard DPO} & \textit{\small (-29\%)} & \textit{\small (+7.6\%)} & \textit{\small (+11\%)} & \textit{\small (-15\%)} & \textit{\small -} & \textit{\small diverse} \\
\bottomrule
\end{tabular}
}
\caption{Results on protein generation. Models are fine-tuned on ProGen2-XL and evaluated on 30K samples. We report sc-RMSD ($\downarrow$), Foldability (pLDDT $> 70$, $\uparrow$), and Energy ($\downarrow$, REU). $\dagger$: ESM-2 uses Gibbs sampling, which is computationally expensive and less comparable to autoregressive models. \textbf{Bold}/\underline{underline}: best/second best.}
\label{tab:main_results}
\vspace{-6pt}
\end{table*}

%% file: tables/zeroshot.tex
\begin{table*}[tbh]
\centering
\resizebox{.95\textwidth}{!}{
\begin{tabular}{l|ccccc|c}
\toprule
\textbf{Method} & \textbf{GFP} & \textbf{GB1} & \textbf{AAV2} & \textbf{TEM-1} & \textbf{P53} & \textbf{Avg.} \\
\textit{(Metric: Spearman's $\rho$)} & \textit{(Stability)} & \textit{(Binding)} & \textit{(Viral)} & \textit{(Resistance)} & \textit{(Suppressor)} & \\
\midrule
\multicolumn{7}{l}{\textit{\textbf{Prior Baselines}}} \\
ESM-2 (3B)~\citep{lin2023evolutionary} & 0.62 & 0.55 & 0.70 & 0.63 & 0.51 & 0.60 \\
Tranception~\citep{notin2022tranception} & 0.65 & \textbf{0.68} & 0.72 & 0.69 & 0.48 & 0.64 \\
\midrule
\multicolumn{7}{l}{\textit{\textbf{ProGen2-XL (6.4B) Backbones}}} \\
ProGen2-XL (Base) & 0.64 & 0.56 & 0.71 & 0.65 & 0.53 & 0.62 \\
SFT & 0.66 & 0.55 & 0.70 & 0.67 & 0.56 & 0.63 \\
PPO (RLHF) & 0.62 & 0.53 & 0.65 & 0.61 & 0.54 & 0.59 \\
DPO (Standard) & \underline{0.71} & 0.59 & \underline{0.73} & \underline{0.70} & \underline{0.61} & \underline{0.67} \\
\midrule
\textbf{Physio-DPO (Ours)} & \textbf{0.78} & \underline{0.63} & \textbf{0.75} & \textbf{0.76} & \textbf{0.70} & \textbf{0.72} \\
\bottomrule
\end{tabular}
}
\caption{Zero-shot fitness prediction on ProteinGym. Spearman correlation ($\rho$) between model log-likelihoods and experimental fitness is reported. All methods use ProGen2-XL. \textbf{Bold}: best result; \underline{Underline}: second best.}
\label{tab:proteingym}
\end{table*}